\documentclass[journal]{IEEEtran}
\ifCLASSINFOpdf
\else
\fi

\usepackage{times}
\usepackage{epsfig}
\usepackage{graphicx}
\usepackage{amsmath}
\usepackage{amssymb}
\usepackage[noend]{algpseudocode}
\usepackage{algorithm}
\usepackage{algorithmicx}
\usepackage{multirow}
\usepackage[table,xcdraw]{xcolor}

\floatname{algorithm}{Algorithm}

\usepackage{cite}
\usepackage[]{hyperref}
\usepackage{pifont}

\begin{document}
%
\title{A Self-Training Approach for Point-Supervised Object Detection and Counting in Crowds}
%
%
%

\author{Yi~Wang, \textit{Graduate Student Member, IEEE}, Junhui~Hou, \textit{Senior Member, IEEE},\\ Xinyu~Hou, \textit{Student Member, IEEE}, and Lap-Pui~Chau, \textit{Fellow, IEEE}
\thanks{Yi~Wang, Xinyu~Hou, and Lap-Pui~Chau are with School of  Electrical and Electronics Engineering, Nanyang Technological University, Singapore, 639798 (e-mail: \{wang1241, houx0008\}@e.ntu.edu.sg, elpchau@ntu.edu.sg).}
\thanks{Junhui~Hou is with the Department of Computer Science, City University of Hong Kong  (e-mail: jh.hou@cityu.edu.hk).}
\thanks{This project was supported in part by Hong Kong Research Grants Council under Grant CityU 11219019 and CityU 11202320.}
\thanks{Corresponding author: Lap-Pui Chau.}
}

%
%

\markboth{IEEE Transactions on Image Processing}%
{Shell \MakeLowercase{\textit{et al.}}: Bare Demo of IEEEtran.cls for IEEE Journals}
%



\maketitle

\begin{abstract}
In this paper, we propose a novel self-training approach named Crowd-SDNet that enables a typical object detector trained only with point-level annotations (i.e., objects are labeled with points) to estimate both the center points and sizes of crowded objects. Specifically, during training, we utilize the available point annotations to supervise the estimation of the center points of objects directly. Based on a locally-uniform distribution assumption, we initialize pseudo object sizes from the point-level supervisory information, which are then leveraged to guide the regression of object sizes via a crowdedness-aware loss. Meanwhile, we propose a confidence and order-aware refinement scheme to continuously refine the initial pseudo object sizes such that the ability of the detector is increasingly boosted to detect and count objects in crowds simultaneously. Moreover, to address extremely crowded scenes, we propose an effective decoding method to improve the detector's representation ability. Experimental results on the WiderFace benchmark show that our approach significantly outperforms state-of-the-art point-supervised methods under both detection and counting tasks, i.e., our method improves the average precision by more than 10\% and reduces the counting error by 31.2\%. Besides, our method obtains the best results on the crowd counting and localization datasets (i.e., ShanghaiTech and NWPU-Crowd) and vehicle counting datasets (i.e., CARPK and PUCPR+)  compared with state-of-the-art counting-by-detection methods. The code will be publicly available at \url{https://github.com/WangyiNTU/Point-supervised-crowd-detection}.

\end{abstract}


%
\IEEEpeerreviewmaketitle
\section{Introduction}

With a massive amount of population living in cities, crowd scenes have become a fundamental yet challenging scenario in a wide variety of computer vision applications, such as video surveillance \cite{SINDAGI2017}, crowd analysis \cite{li2015crowded,kang2017beyond}, and safety monitoring \cite{chan2008privacy,chen2011real}. Objects in dense crowds present small sizes, large scale variations, and high occlusions, which poses great challenges to object detection methods that simultaneously predict objects' locations and sizes in an image. 

\begin{figure} [t]
\centering 
\includegraphics[width=9cm]{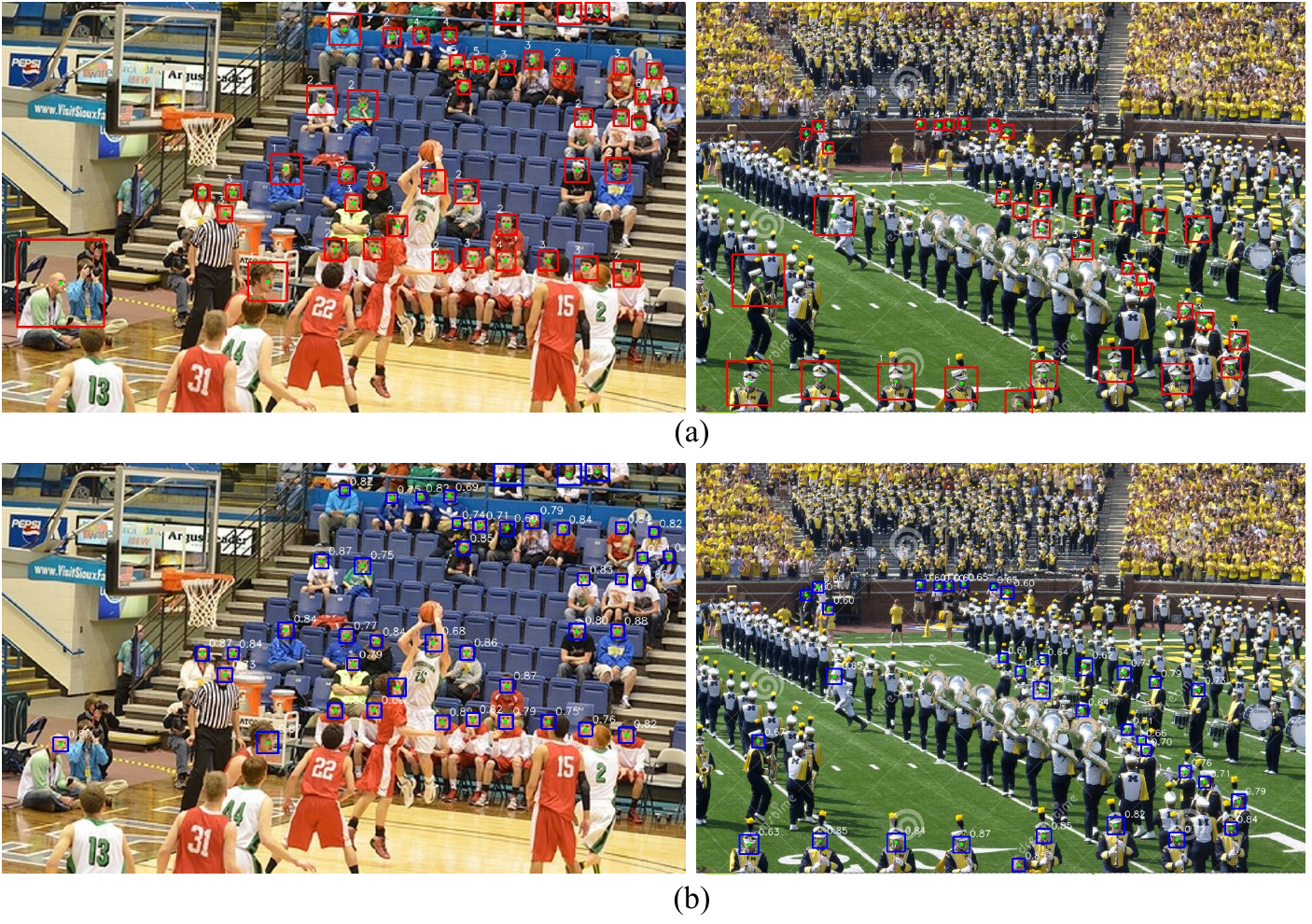}
\caption{Illustration of generated training examples at different phases by our self-training approach. The images are from WiderFace dataset \cite{yang2016wider}. The green dots shown at the centers of faces stand for the point-level annotations. (a): Before training, the pseudo object sizes (red boxes) generated by the proposed locally-uniform distribution assumption. The numbers shown at the top-right corner of red boxes stand for object crowdedness. (b): After training, the refined pseudo object sizes (blue boxes) by our crowdedness-aware loss and the confidence and order-aware refinement scheme. The numbers shown at the top-left corner of blue boxes stand for the pseudo object sizes' posterior probabilities. \textit{Zoom in the figure for better viewing}.}
\label{fig1}
\end{figure}

The advances of deep neural networks (DNNs) raise an issue of enormous demand for data annotations. However, it is very costly and laborious to collect object-level bounding box annotations \cite{lin2014microsoft_coco,wang20000pedestrian_benchmark} which are usually needed for training DNN-based object detection methods, especially for images containing thousands of objects. Current crowd counting datasets provide only point-level annotations, and usually human heads are labeled as the central points, e.g., the green dots shown in Fig. \ref{fig1}. Due to the lack of object sizes, state-of-the-art DNN-based object detectors \cite{zhao2019object} cannot be trivially applied to such point supervision. As pioneers, Liu \textit{et al.} \cite{liu2019point} introduced a pseudo size updating scheme in a detection network to estimate object sizes. Sam \textit{et al.} \cite{sam2019locate} proposed an LSC-CNN to achieve higher detection performance in crowd scenes. However, these works are still not on par with box-supervised methods (e.g., Faster R-CNN \cite{ren2015faster}) in the detection task. As for the counting task, these methods, denoted as counting-by-detection methods, can count objects by filtering out low-confidence objects with a threshold. However, they also suffer from highly crowded objects.

Alternatively, modern crowd counting methods \cite{li2018csrnet,shi2019counting,gao2020cnn}, named counting-by-regression methods, bypass the locations and sizes of objects but employ DNNs to regress a density map, which is further integrated to obtain the overall count. These methods have achieved outstanding counting performance in dense crowds. However, they only aim to count the number of objects and lose individual information, i.e., object instance's location and size. We argue that the density map only contains weak information about the crowds, while locations and sizes of object instances provide much more important information for other computer vision applications, such as multi-object tracking \cite{rodriguez2011density,xiang2015learning}, face recognition \cite{turk1991face,yang2016wider}, and person re-identification \cite{li2014deepreid}.

In view of these issues, we propose a novel self-training approach capable of training a typical detection method only with point-level annotations such that it can accurately and simultaneously detect and count objects in dense crowds. Specifically, based on a keypoint-based detector, i.e., center and scale prediction (CSP) \cite{Liu_2019_CVPR}, we decouple detection as the separate estimation of objects' center points and sizes. The available point-level annotations directly supervise the estimation of the center points during training. As the ground-truth object sizes are not accessible, we propose a simple yet effective assumption in crowd scenes, called locally-uniform distribution assumption (LUDA), to generate the initial pseudo size for each object (see the red bounding boxes shown in Fig. \ref{fig1}(a)). Meanwhile, we propose a crowdedness-aware loss to emphasize the contributions of crowded objects in object size regression (see the crowdedness at the top-left corner of red boxes in Fig. \ref{fig1}(a)). Moreover, we propose a confidence and order-aware refinement scheme to continuously update the pseudo sizes during training, which performs the refinement operation by considering both the prior confidences and the updating order of pseudo sizes, such that the detection ability of the detector is increasingly boosted (see the blue bounding boxes shown in Fig. \ref{fig1}(b) for the refined pseudo object sizes). Besides, to deal with highly dense crowds (e.g., one person represented by several pixels in an image on the ShanghaiTech \cite{Zhang2016Single} dataset), we propose an effective decoding method to improve the representation ability of the detector, in which a feature fusion and decoding technique is employed to restore the full-resolution feature maps. 

Extensive experimental results show that our approach outperform start-of-the-art point-supervised methods to a significant extent in terms of the detection performance, i.e., more than 10\% AP improvement is achieved. Moreover, our method even produces comparable performance to the box-supervised Faster R-CNN on the WiderFace benchmark \cite{yang2016wider}. For center point localization, our approach produces the best results among state-of-the-art methods on dense crowd datasets, e.g., ShanghaiTech \cite{Zhang2016Single} and NWPU-Crowd \cite{wang2020nwpu}. For crowd counting, our method obtains comparable results to the latest counting-by-regression methods. Note that the bounding boxes produced by our method are more informative than the density map produced by counting-by-regression methods. 

The rest of this paper is organized as follows. In Section \ref{Sec:Related}, we introduce the related works on object detection and counting in crowds. Then, the proposed self-training approach is presented in detail in Section \ref{Sec:Method}. Experimental results and ablation studies are provided in Section \ref{Sec:Exp}. Finally, we conclude this paper in Section \ref{Sec:Con}.

\begin{figure*} [ht]
\centering 
\includegraphics[width=\textwidth]{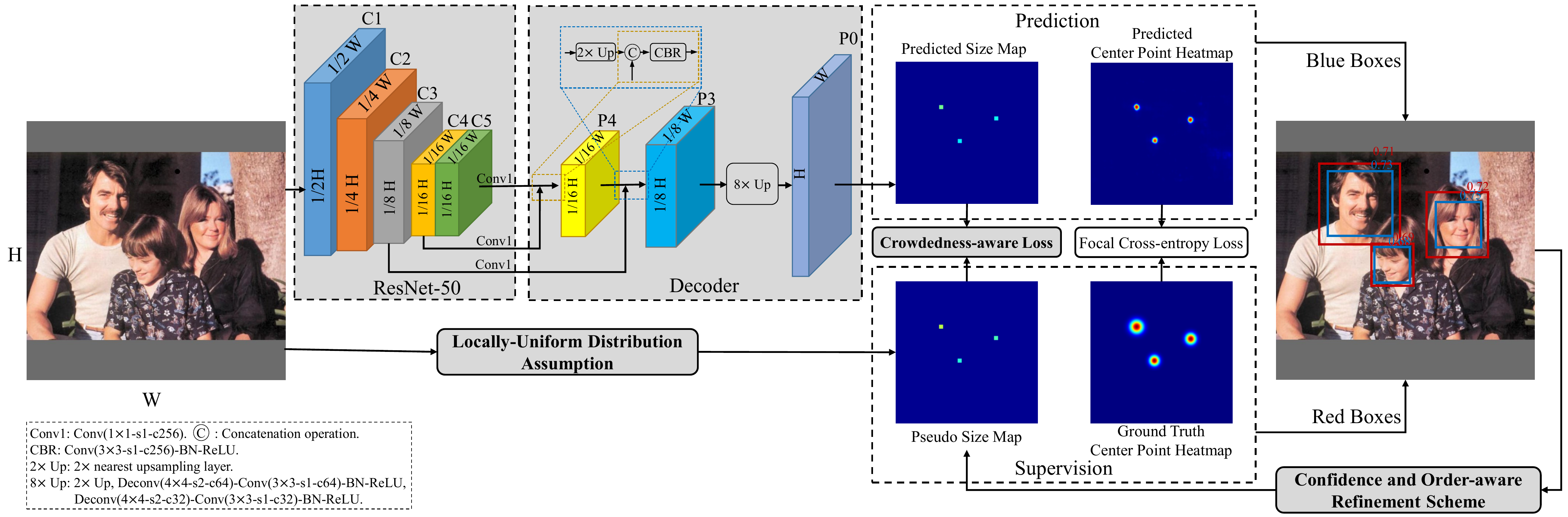}\\ 
\caption{Illustration of the proposed self-training framework for training a detector only with point-wise supervision. The decoder produces high-resolution feature maps for the estimation of objects' center points and sizes. Before training, the pseudo object sizes are first generated from the point-wise supervisory information, based on our locally-uniform distribution assumption. During training, the pseudo object sizes are further refined by our confidence and order-aware refinement scheme under the supervision of the proposed crowdedness-aware loss. Here an image with sparse objects is used for clear visualization purposes. }\label{fig2} 
\end{figure*}

\section{Related Work}
\label{Sec:Related}
\subsection{Object Detection in Dense Crowds}
Benefiting from the advances of DNNs, recent object detectors, such as Faster R-CNN \cite{ren2015faster}, RetinaNet \cite{Lin_2017_ICCV}, and CenterNet \cite{zhou2019objects,duan2019centernet}, achieve appealing performance. Despite the remarkable progress made, these methods encounter difficulties when counting small and heavily occluded objects in crowded scenes. To enhance the detectors' abilities, Liu \textit{et al.} \cite{Liu_2019_CVPR} proposed a keypoint-based detector named CSP to predict the central points and scale of objects separately. Goldman \textit{et al.} \cite{goldman2019precise} presented a deep-learning-based method for precise object detection in densely packed scenes. They introduced a soft intersection-over-union (IoU) network layer and an expectation-maximization (EM) based clustering method to deal with overlapped objects in the dense scenes. Though these above detectors all achieve good performance, they have to be trained and supervised by box-annotated examples.

Most crowd counting datasets only provide point annotations for denser crowds. It is nontrivial to train a detector with point supervision. Similarly, sample weighting  techniques \cite{cai2020learning_weighting} that require the box supervision fail to work on point-supervised detection. Recently, several works begin to use the blobs \cite{laradji2018blobs} to localize the individuals in crowds and even to estimate the sizes of the heads \cite{liu2019point,sam2019locate}, with only point-level annotations. Laradji \textit{et al.} \cite{laradji2018blobs} argued that the unnecessary size and shape information drags the performance of detection-based methods in counting problems. A localization-based counting loss that combines image-level, point-level, split-level, and false-positive loss is used to train a fully convolutional network (FCN) such that it could produce the blobs in the center of objects. Based on a regression-based network, Idrees \textit{at al.} \cite{idrees2018composition} proposed a post-processing method to find the local peaks on the density map as the center locations of heads. As a baseline method, Sam \textit{et al.} \cite{sam2019locate} applied a threshold technique on the density map of the CSRNet \cite{li2018csrnet} to obtain detections (called CSR-A-thr). However, these methods only localize the center points of individuals in crowds.

To further estimate individuals' sizes, Liu \textit{et al.} \cite{liu2019point} proposed a detection network, named PSDDN, which builds a strong baseline for point-supervised detection and counting in crowds. The PSDDN employs the nearest neighbor distance \cite{Zhang2016Single} to initialize the pseudo boxes and updates the pseudo boxes by choosing smaller box predictions. Another state-of-the-art method, named LSC-CNN, was proposed recently by Sam \textit{et al.} \cite{sam2019locate}, `
\subsection{Object Counting in Dense Crowds}
Bypassing localization, counting-by-regression methods \cite{SINDAGI2017,Zhang2016Single,zhang2015cross,wang2019object} were proposed to address the crowd counting problem and have dominated this field for years. Instead of directly regressing the global count adopted in early works \cite{chan2008privacy,chan2009bayesian}, current approaches \cite{li2018csrnet,shi2019counting,cao2018scale} exploit DNNs to estimate a density map \cite{lempitsky2010learning}, over which the count is obtained via the integration operation. The typical network architecture of this kind of methods is generally composed of an encoder for extracting a set of features from an image and a decoder for regressing a density map \cite{DensityAware}. For example, in \cite{Zhang2016Single,zhang2015cross}, a multi-column structure was introduced to learn the multi-scale features for representing large scale variation of objects. Li \textit{et al.} \cite{li2018csrnet} demonstrated the redundancy of features in the multi-column structure and proposed a single-column deep structure (e.g., VGG \cite{simonyan2014very}) with dilated convolution layers, which achieved better results. Cao \textit{et al.} \cite{cao2018scale} proposed a scale aggregation module to learn the scale diversity of features. Shi \textit{et al.} \cite{shi2019counting} repurposed point annotations as a segmentation map and a global density map. 

By exploiting the attention mechanism, Sindagi \textit{et. al.} \cite{sindagi2019ha} proposed an attention-based crowd counting network, named HA-CCN, to enhance the different-level features of the network. To realize a lightweight model, Wang \textit{et al.} \cite{DensityAware} proposed a pixel shuffle decoder (PSD) to generate a high-resolution density map without using convolutional layers. They presented a density-aware curriculum learning (DCL) strategy to improve the network's generalization ability and reduce training time. Recently, Wang \textit{et al.} \cite{wang2020nwpu} provided a large-scale benchmark, namely NWPU-Crowd, for the crowd localization and counting. A recent survey paper written by Gao \textit{et al.} \cite{gao2020cnn} reviews over two-hundred crowd counting works and some datasets, showing the improvement in this filed.

Although the counting-by-regression methods obtain state-of-the-art counting performance, they sacrifice the location and size information of objects \cite{gao2020cnn}. We argue that the count is rough information for the crowds and not enough for further use of high-level vision tasks. In our work, we count objects by our object detections.

\subsection{Counting from Drone View}
Another dense case appears in vehicle counting from drone images. Hsieh \textit{et al.} \cite{hsieh2017drone} introduced a large-scale car parking lot dataset (CAPPK), which consists of approximately 90k cars. Motivated by the regular spatial layout of cars, they proposed a layout proposal network (LPN) to count vehicles. Without local annotations such as bounding boxes or points, Stahl \textit{et al.} \cite{stahl2018divide} proposed a general object counting method that uses local image regions to predict the global image-level counts. Goldman et al. \cite{goldman2019precise} considered that vehicles in drone images are a densely packed scene. Therefore, they used their object detection method in this scenario. To deal with the failure of state-of-the-art detectors in drone scenes, Li \textit{et al.} \cite{li2019simultaneously} made a set of modifications for detection. An effective loss is proposed to yield the scale-adaptive anchors. Then, the circular flow is applied to guide feature extraction. Third, a counting regularized constraint is introduced to the loss function. 

\section{Proposed Method}
\label{Sec:Method} 
Fig. \ref{fig2} illustrates the proposed self-training framework, which is capable of training a typical object detector only with point-level annotations for simultaneous object detection and counting. To be specific, the framework is based on an anchor-free keypoint-based object detector, i.e., CSP detector \cite{Liu_2019_CVPR}. A locally-uniform distribution assumption is proposed to generate the initial pseudo size for each object, and a crowdedness-aware loss is proposed to emphasize the pseudo sizes of crowded objects in size regression. Furthermore, a confidence and order-aware refinement scheme is proposed to update the pseudo sizes in each training iteration. In addition, a decoding method is proposed to handle dense crowds. In what follows, we present the detection network and the self-training approach in detail. 

\subsection{Detection Network}
\label{subsec:detectionnet}
We employ the keypoint-based detection method because it allows us to estimate the center point and size of objects separately. Therefore, the center points could be directly supervised by the point-level annotation, while the size estimation is accomplished by the proposed modules. In addition, it is anchor box-free. 

\subsubsection{Architecture}
For the datasets with medium-density crowds (e.g., WideFace \cite{yang2016wider}, CARPK and PUCPR+ \cite{hsieh2017drone}), we employ the original network in \cite{Liu_2019_CVPR}. For the dataset with high-density crowds (e.g., ShanghaiTech \cite{Zhang2016Single} and NWPU-Crowd \cite{wang2020nwpu}), however, the original network performs poorly since the center point prediction fails to represent highly dense objects in its output feature map of size \(\frac{H}{4}\times \frac{W}{4}\), where \(H\) and \(W\) are the height and width of the input image, respectively. Hence, we propose an effective decoding method to handle this issue. 

As illustrated in Fig. \ref{fig2}, the detection network contained in our approach adopts the five-stage ResNet-50 \cite{he2016deep} as the backbone network, where each stage downsamples the feature maps by a factor of 2, except for Stage-5 that uses the dilated convolutions to keep the stride the same as Stage-4. Let \(C_{i}, i \in \{1,2,...,5 \}\) be the output feature maps of the $i$-th stage. In the \textit{Decoder}, three Conv(1$\times $1-s1-c256) are applied to reduce the number of channels of \(C_{5}\), \(C_{4}\), and \(C_{3}\), where Conv denotes the convolutional layer, and (1$\times $1-s1-c256) means the layer with the kernel of 1$\times $1, the stride of 1, and the channel of 256. Then, we employ a top-down feature fusion manner to merge \(C_{5}\), \(C_{4}\), and \(C_{3}\), generating the fused features \(P_{3}\). The fusion manners are shown as the dotted yellow and blue rectangle in Fig. \ref{fig2}. Finally, we use an 8$\times$ Up structure to decode the fused features, consisting a 2$\times$ nearest upsampling layer followed by two Deconv(4$\times $4-s2)-Conv(3$\times $3-s1)-BN-ReLU, where Deconv, BN, and ReLU denote the deconvolution, batch normalization, rectified linear unit, respectively. The decoder produces the output feature maps (\(P_{0}\)) with the same size as the input, i.e., \(H\times W\). There are two separate heads (i.e., Conv(1$\times $1-s1-c1)) for center point prediction and size prediction, producing the center point heatmap and the size map. 

\subsubsection{Supervision information}
Let $\{\mathbf{p}_j\}_{j=1}^M$ be the point-wise annotations, where \(p_{j} := (x_{j}, y_{j})\) is the 2D coordinates of the center of the \(j\)-th object in an image, and \(M\) is the total number of objects in an image. As shown in Fig. \ref{fig2}, to supervise the estimation of object center points, we generate a ground-truth center point heatmap \(Q \in [0, 1]^{H\times W}\) with \(1\) for objects' center points and \(0\) for negative points. To decrease the ambiguity of negative points surrounding the positive ones, we place a normalized 2D Gaussian mask at the center location of each positive point, as performed in \cite{Liu_2019_CVPR}. If two masks overlap, we choose the element-wise maximum for the overlapped region. For the object size supervision, we generate pseudo object sizes denoted by \(s_{j}\), which will be introduced in Sec. \ref{subsec:sizegen}. Here we assume that the objects (e.g., heads and faces) have an aspect ratio of \(1\) in crowded scenes. We assign \(log(s_{j})\) to the \(j\)-th object's center coordinates \((x_{j}, y_{j})\) and zeros to other locations, generating the size map \(S \in \mathbb{R}^{H\times W}\). For the original CSP with the output of size \(\frac{H}{4}\times \frac{W}{4}\), an offset map is appended to estimate the discretization error caused by the stride of 4. The ground-truth offset for \(p_{j}\) is defined as \(\frac{x_{j}}{4} - \left \lfloor \frac{x_{j}}{4} \right \rfloor\) and \(\frac{y_{j}}{4} - \left \lfloor \frac{y_{j}}{4} \right \rfloor\) on the $x$-axis and $y$-axis, respectively, which is assigned to \((x_{j}, y_{j})\) on the offset map. \(\lfloor r\rfloor: \mathbb{R} \to \mathbb{Z}\) of a real number \(r\) denotes the greatest integer less than or equal to \(r\). 

\subsubsection{Loss function}
We apply the focal cross-entropy loss \cite{Liu_2019_CVPR,zhou2019objects} to each pixel on the center point heatmap: 
\begin{equation}
L_{c}=-\frac{1}{M}\sum_{j=1}^{M}\begin{cases}
(1-\hat{q}_{j})^{\gamma }log(\hat{q}_{j}), & \text{ if } q=1, \\ 
A(1-q_{j})^{\delta}(\hat{q}_{j})^{\gamma }log(1-\hat{q}_{j}), & \textrm{otherwise}, 
\end{cases}
\end{equation}
where \(\hat{q}_{j}\) and \(q_{j}\) are the predicted probability and the ground-truth label of pixel $j$, respectively; \(\gamma\) is the hyper-parameter of the focal loss \cite{Lin_2017_ICCV} which is set to 2 in all experiments; \(\delta\) is the hyper-parameter to control the penalty of negatives, which is set to 4 in all experiments; \(A\) is the coefficient to address the imbalance between positive and negative points, which is set to 1 (resp. 1/16) for the original CSP (resp. the proposed decoding structure) in all experiments. Intuitively, if we enlarge the CSP's output feature map by the decoding method from \(\frac{H}{4}\times \frac{W}{4}\) to \(H\times W\), the number of negative points will increase by 16 times. Hence, we set \(A=1/16\) to balance the positive and negative points. For object size regression, we propose a crowdedness-aware loss \(L_{size-\alpha}\), of which the details will be described in Sec. \ref{subsec:reweight}. For offset estimation, the smooth L1 loss \cite{Girshick2015Fast} is calculated between the ground-truth offsets and predicted ones, denoted as \(L_{o}\). The overall training objective is
\begin{equation}
L = \lambda L_{c} + L_{size-\alpha} + L_{o},
\end{equation}
where \(\lambda\) is the weight for center point classification, which is experimentally set to \(0.1\).

\subsubsection{Inference}
The detector first performs a forward pass to generate the center point heatmap, the size map, and the offset map (if it is used). The peak points in the center point heatmap are extracted by a 3$\times$3 max pooling operation. Then, we obtain the center point coordinates of objects whose probabilities are larger than a predefined confidence. The object sizes are obtained from the corresponding coordinates in the size map (see the ``Prediction'' in Fig. \ref{fig2}). If the offset map is appended, the corresponding offsets are added to object coordinates. Finally, the bounding boxes can be decoded by the coordinates and sizes.

\subsection{LUDA-based Pseudo Object Size Generation}
\label{subsec:sizegen}
As the ground-truth object sizes are unavailable, we generate pseudo object sizes from the point-wise supervisory information to train the detector. In crowded scenarios, object instances, e.g., heads, faces, or cars, are usually uniformly distributed in an image. According to the geometry-adaptive kernel \cite{Zhang2016Single}, a typical object's size is proportional to the distance to its nearest neighbors in dense crowds. This assumption is relatively weak as the objects are not always dense enough in an image. In this paper, we employ the non-uniform kernel \cite{shi2019counting} to locally restrict the above assumption  and propose the locally-uniform distribution assumption (LUDA). That is, 1) the objects in crowd scenarios are uniformly distributed in a local region and have a similar size in that region; and 2) the crowdedness of the region affects the precision of size estimates. In what follows, we detail our pseudo object size generation method based on LUDA. 

Following \cite{Zhang2016Single,shi2019counting}, we first calculate the initial object size of point \(\mathbf{p}_{j}\) according to the distances to its $K$ nearest points, i.e., 
\begin{equation}
\label{eqn1}
\overline{d}_{j}=\frac{1}{K}\sum_{k=1}^{K} \beta d_{j,k}, 
\end{equation}
where \(\overline{d}_{j}\) is the initial object size of \(\mathbf{p}_{j}\), \(d_{j,k}\) is the distance between point \(\mathbf{p}_{j}\) and its \(k\)-th nearest neighbor, and \(\beta\) is a scalar. The initial object size is further smoothed to reduce the variation in a local region, leading to the pseudo object size \(s_{j}\):
\begin{equation}
\label{eqn2}
s_{j}=\frac{1}{\left | R_{p_{j}} \right |}\sum_{l\in R_{p_{j}}} \overline{d}_{l} ,
\end{equation}
where \(R_{p_{j}}\) is the \(p_{j}\)-centered local region, \(\left | R_{p_{j}} \right |\) is the number of the points inside \(R_{p_{j}}\), and \(\overline{d}_{l}\) is the initial size of the \(l\)-th point contained in \(R_{p_{j}}\). We set the \(R_{p_{j}}\) to a circular region so that KD-Tree can be used to improve the calculation speed. \(\left | R_{p_{j}} \right |\) affects the precision of size estimates, which will be further exploited for the crowdedness-aware loss in Sec. \ref{subsec:reweight}.

Since the crowdedness differs from regions to regions, a single set of parameters in Eqs. (\ref{eqn1}) and (\ref{eqn2}) can only fit a specific crowdedness. In experiments, we choose multiple sets of parameters intuitively for different training sets such that the crowded objects have precise pseudo sizes. The detailed parameter settings will be explained in Sec. \ref{subsubsec:implement}. 

\subsection{Crowdedness-aware Loss for Object Size Regression}
\label{subsec:reweight}
With the pseudo object size $s_j$, a straightforward way to supervise the regression of object sizes is using the smooth L1 loss \cite{Girshick2015Fast}, i.e., 
\begin{equation}
\label{eqn3}
L_{s} = \frac{1}{M} \sum_{j=1}^{M} SmoothL1(\hat{s}_{j}, s_{j}),
\end{equation}
where \(\widehat{s}_{j}\) is the size prediction of the \(j\)-th object. However, there is an obvious drawback for Eq. (\ref{eqn3}), i.e., all object instances equally contribute to the loss, and if some pseudo object sizes are inaccurate (i.e., noisy supervision), the training of the detector will be adversely affected. Moreover, inaccurate pseudo sizes are inevitable since the simple LUDA cannot resolve the complexity of crowded scenes (see Fig. \ref{fig1}(a)). Such a drawback is experimentally verified in the following Table \ref{tabel2} (see the 5th entry of Table \ref{tabel2}). 

To this end, we propose a crowdedness-aware loss associating each object instance with its crowdedness, which indicates the importance in object size regression. We formulate the crowdedness-aware loss as
\begin{equation}
\label{eqn4}
L_{size-\alpha}=\frac{1}{M}\sum_{j=1}^{M} \alpha_{j} SmoothL1( \widehat{s}_{j}-s_{j}),
\end{equation}
where \(\alpha_{j} \in [0,+\infty)\) is the crowdedness-aware factor controlling weight of the \(s_{j}\). Based on LUDA, crowded objects can produce more accurate pseudo object sizes than sparse ones, and thus the crowded objects should be assigned with larger weights. Specifically, we define the \(\alpha_{j}\) as an exponential function of the number of objects (crowdedness) inside the local region \(R_{p_{j}}\), i.e.,
\begin{equation}
\label{eqn5}
\alpha_{j}=(\left | R_{p_{j}} \right |)^{\eta}.
\end{equation}
Here, we use a tunable parameter \(\eta \geq 0\) to scale the factor. In practice, we use a threshold to limit the maximum value of \(\alpha_{j}\) to avoid gradient explosion. By assigning crowdedness-aware factors to the pseudo sizes, the loss function emphasizes the influence of the crowded objects and weakens that of the sparse ones. The crowdedness-aware loss enhances the robustness of the training with the noisy pseudo sizes.

\subsection{Confidence and Order-aware Refinement Scheme for Pseudo Size Updating}

Although the crowdedness-aware loss is able to boost the robustness of the detector, the inaccurate sizes still exist and will affect the backward pass of training. Thus, we propose a confidence and order-aware refinement scheme to update pseudo object sizes for better training the detector. In \cite{liu2019point}, the pseudo bounding boxes are updated by selecting the predicted boxes with the highest scores among those whose sizes are smaller than the pseudo ones. However, such a criterion may not be true in practice since it ignores the following two key issues, resulting in the inaccurate refinement and unstable training process. 1) The prior information. It updates the pseudo sizes without considering their prior confidences. 2) The updating order. All pseudo sizes are treated identically, resulting in the synchronous updating of both accurate and inaccurate sizes.

Instead, by taking prior information into account, we assign every pseudo size with a prior probability at the beginning of training, and if and only if the predicted posterior probability of the detector is larger than the prior probability, we update the pseudo size (resp. prior probability) with the predicted size (resp. posterior probability) for the next epoch. Referring to the prior confidence, our refinement scheme guarantees the detector is trained with increasingly confident examples. Meanwhile, it can update the most inaccurate sizes first, then followed by updating the relatively accurate ones. This is achieved by setting the same prior probability for all pseudo sizes before training. During training, easy examples (e.g., sparse and large objects) with noisy size supervision that achieve high posterior probabilities rapidly are updated first. Then, hard examples (e.g., crowded and small objects) with more accurate size supervision are updated. In other words, the updating order is from noisy sizes to noiseless sizes. The proposed refinement scheme presents a more powerful error-correction capability than the one in \cite{liu2019point}. See the experimental results in Table \ref{tabel2}.

\begin{algorithm}[t]
\footnotesize
\caption{Confidence and Order-aware Refinement Scheme.}\label{lctf}
\begin{algorithmic}[1]
\Require The \(i\)-th input image $X_i$ with a set of pseudo bounding boxes $B_{i}=\{b_{1},...,b_{M_{i}}\}$, and prior probabilities of the boxes \(P(b_{j}), j \in \{1,...,M_{i}\}\).
\Ensure Refined pseudo bounding boxes $\widetilde{B}_{i}=\{\widetilde{b}_{1},...,\widetilde{b}_{M_{i}}\}$, and refined prior probabilities \(P(\widetilde{b}_{j})\).
\vspace{0.3cm}
\State Forward passing the detector to generate the detections $\widehat{B}_{i}=\{\widehat{b}_{1},...,\widehat{b}_{M_{i}}\}$ with the posterior probabilities \(P(C, \widehat{B}_{i}|X_{i})\).
  \For {$j$ in $\{1,...,M_{i}\}$} 
      \If {$P(C, \widehat{b}_{j}|X_{i}) > P(b_{j})$} 
          \State $P(\widetilde{b}_{j}) \leftarrow P(C, \widehat{b}_{j}|X_{i})$
          \State $\widetilde{b}_{j} \leftarrow \widehat{b}_{j}$
      \Else
          \State $P(\widetilde{b}_{j}) \leftarrow P(b_{j})$
          \State $\widetilde{b}_{j} \leftarrow b_{j}$
      \EndIf
  \EndFor
\end{algorithmic}
\label{algo1}
\end{algorithm}

More specifically, let \(b_{j}= \left \{p_{j},  s_{j}\right \}\) be a pseudo bounding box centered at $p_{j}$ with the pseudo size $s_{j}$, and the prior probability $P(b_{j})$ is assigned to $b_{j}$. The object detection problem is cast as learning the posterior \(P(C, B|X)\), where \(X\) is the input image, \(B\) is the bounding boxes of objects, and \(C \in \left \{0,  1\right \}\) is the binary class with 0 for background and 1 for object instance. Our refinement scheme, which is merged into the training process, is summarized in Algorithm \ref{algo1}. In each training iteration, after executing a forward path, we obtain the predicted detections $\widehat{B}_{i}$ with the posterior \(P(C, \widehat{B}_{i}|X_{i})\) (i.e., Line 1). For each instance \(j \in \{1,...,M_{i}\}\), if $P(C, \widehat{b}_{j}|X_{i})$ is larger than $P(b_{j})$, we update the prior probabilities \(P(\widetilde{b}_{j})\) with the posterior \(P(C, \widehat{b}_{j}|X_{i})\), and update the pseudo bounding boxes \(\widetilde{b}_{j}\) with the prediction \(\widehat{b}_{j}\) (i.e., Lines 3-5). Otherwise, the prior probability and pseudo bounding box remain unchanged (i.e. Lines 7 and 8). A constant prior probability of \(0.6\) is assigned to all $\{B_{i}\}_{i}^{N}$ at the beginning of training, where \(N\) is the number of training images.
 
\textit{Remark}. Here we provide more explanations why our self-training approach enables the detector to update the pseudo size towards the correct direction. The main reasons come from two aspects. First, as analyzed in  Sec. \ref{subsec:sizegen}, our LUDA-based pseudo size generation method ensures that the pseudo sizes of crowded objects (i.e., hard examples) are more accurate than those of sparse objects. Also, it is known that hard examples have a heavy influence on the detector's training than easy examples \cite{shrivastava2016training,roychowdhury2019automatic}. Thus, when incorporated with the crowdedness-aware loss, the larger number of relatively accurate hard examples in crowded scenes dominate the detector, making it robust to the outliers (e.g., inaccurate pseudo sizes). Second, the pseudo sizes are updated in a robust and orderly manner by our refinement scheme. We set the same initial prior probability for all pseudo bounding boxes. The posterior probability of easy examples will first meet the initial prior probability during training, and thus Algorithm \ref{algo1} begins with updating the pseudo sizes of inaccurate easy examples. With increasingly accurate easy examples, the detector becomes stronger such that hard examples are then refined. More experimental results to illustrate the effectiveness of the proposed method can be found in Sec. \ref{subsec:ExpAbl}.

\section{Experimental Results}
\label{Sec:Exp}
In this section, we first describe the experiment settings, including datasets, implementation details, and evaluation metrics, followed by ablation studies for verifying the effectiveness of each component of our approach. Finally, we compare our approach with state-of-the-art methods in terms of both detection and counting tasks.

\subsection{Experiment Settings}
\label{subsec:ExpSetting}

\subsubsection{Datasets}
We used five representative datasets in crowd scenes, i.e., WiderFace \cite{yang2016wider} for dense face detection, ShanghaiTech \cite{Zhang2016Single} for crowd counting and localization, NWPU-Crowd \cite{wang2020nwpu} for crowd localization, and CARPK and PUCPR+ \cite{hsieh2017drone} for vehicle counting from the drone view.

\textbf{WiderFace} is one of the most challenging face detection benchmarks, where the 32,203 images contain 393,703 human-labeled faces were captured in a wide variety of imaging conditions, such as large variations in scale and pose, high occlusion, and changeable illumination conditions. 40\%, 10\%, and 50\% of the images were used for training, validation, and testing, respectively. Following existing point-supervised detection methods \cite{sam2019locate,liu2019point}, we trained our model on the training set, and reported detection and counting results on the validation set. 

\textbf{ShanghaiTech} presents high-density crowds, which contains 482 images on Part\_A (SHA) and 716 images on Part\_B (SHB). The number of people in an image ranges from 33 to 3139 on SHA, and 9 to 578 on SHB. We followed the training and testing split in \cite{Zhang2016Single} to evaluate the counting and central point localization performance. 

\textbf{NWPU-Crowd} provides a large-scale benchmark for crowd counting and localization. It consists of 3109, 500, and 1500 images for the training, validation, and test, respectively, and the images contain more than 2 million annotated heads with points and boxes. 

\textbf{CARPK and PUCPR+} are composed of images of parking lots from the drone view and high-rise buildings, respectively. CARPK contains nearly 90k cars, while PUCPR+ contains about 17k cars in total. We used the evaluation protocol in their benchmark to evaluate the counting performance of our method.

\begin{table}[t]
\centering \caption{The parameter settings of the LUDA-based pseudo size generation method for different datasets. ``-'' means the parameter is not required.}
\resizebox{\columnwidth}{!}{%
\begin{tabular}{l|c|c|c}
\hline
\multirow{2}{*}{Dataset} & \multicolumn{3}{c}{Generation Parameters} \\ \cline{2-4} 
      & $K$ of Eq. (\ref{eqn1})    & $\beta$ of Eq. (\ref{eqn1})    & Max $\alpha_{j}$  of Eq. (\ref{eqn5})  \\ \hline
WiderFace \cite{yang2016wider}   & 2    & 0.5      & -        \\
SHA \cite{Zhang2016Single}     & 2       & 1          & 50                  \\
SHA \cite{Zhang2016Single}    & 2       & 0.5        & 50                  \\
NWPU-Crowd \cite{wang2020nwpu}    & 2       & 0.8        & 200               \\
CARPK \cite{hsieh2017drone}  & 1       & 1.3        & -                \\
PUCPR+ \cite{hsieh2017drone}  & 1       & 1.2        & -                \\ \hline
\end{tabular}
}
\label{tabel0}
\end{table}

\subsubsection{Implementation details}
\label{subsubsec:implement}
The backbone, ResNet-50, was initialized with the pre-trained weights on ImageNet \cite{krizhevsky2012imagenet}. We trained our detector on 3 GPUs with the batch size of 12. We adopted Adam \cite{kingma2014adam} optimizer with the learning rate of \(7.5 \times 10^{-6}\) for the WiderFace dataset and \(7.5 \times 10^{-5}\) for the remaining ones.  The input images were randomly re-scaled, color distorted, flipped, and then cropped into $704\times 704$ image patches. We used the same re-scale technique as CSP \cite{Liu_2019_CVPR}. All training processes can only access to the point annotations. For the datasets with bounding box annotations, i.e., WiderFace, CARPK, and PUCPR+, we calculated their center points for training. We stopped training at 200k, 8k, 60k, 45k, and 4.5k iterations for WiderFace, ShanghaiTech, NWPU-Crowd, CARPK, and PUCPR+, respectively. Like \cite{Liu_2019_CVPR,liu2019point}, we performed the multi-scale testing to generate bounding boxes. Then, non-maximum suppression (NMS) was used to filter the generated boxes. Following \cite{sam2019locate}, we randomly took 10\% training images of ShanghaiTech as the validation set. For ShanghaiTech and NWPU-Crowd, we chose the best model by performing a threshold search to minimize the counting error on the validation set. The other datasets adopted the confidence threshold of 0.4 to produce boxes for counting.

As mentioned in Sec. \ref{subsec:sizegen}, we experimentally chose multiple sets of parameters by which the precise pseudo sizes were generated for crowded objects on training sets. Table \ref{tabel0} shows the parameter settings used in Eqs. (\ref{eqn1}) and (\ref{eqn5}) for different datasets. We especially set the maximum value of $\alpha_{j}$ to 50 to avoid large gradients since dense crowds of ShanghaiTech produce large values of $\alpha_{j}$. For all datasets, we set \(\eta=1\) in Eq. (\ref{eqn5}). 

\begin{table}[t]
\centering \caption{Comparisons of our LUDA-based pseudo size generation method with the GAK-based method on the training set of WiderFace \cite{yang2016wider}. Refinement of pseudo object sizes by our self-training approach is listed at the last entry. The number in the parentheses denotes the IoU threshold for AP calculation. The larger the value of AP is, the better.}
\begin{tabular}{l|ccc}
\hline
\hline
Size Generation & AP (0.3)      & AP (0.5)      & AP (0.7)      \\ \hline
GAK \cite{Zhang2016Single} & 55.8          & 29.5          & 7.0           \\
LUDA (ours)    & 60.3          & 31.2          & 7.5           \\
Refined size (ours)  & \textbf{79.4} & \textbf{50.6} & \textbf{14.4} \\ \hline
\end{tabular}
\vspace{0.2cm}
\label{tabel1}
\end{table}

\begin{table*}[t]
\centering \caption{The ablative studies towards the crowdedness-aware loss and the confidence and order-aware refinement scheme, as well as comparisons with state-of-the-art point-supervised detection method \cite{liu2019point} and counting method \cite{shi2019counting} on the validation set of WiderFace \cite{yang2016wider}. ``C-by-D'' means counting by detection, and ``C-by-R'' means counting by regression. ``Size Init.'' stands for directly applying the pseudo object size initialization method on the validation set to obtain the detection result. ``-'' is not applicable. The larger the value of AP is, the better. The lower the values of MAE and NAE are, the better.}
\begin{tabular}{l|c|cc|ccc|cc}
\hline
                                           &            &   &            & \multicolumn{3}{c|}{Detection AP} & \multicolumn{2}{c}{Counting} \\ \cline{5-9} 
\multirow{-2}{*}{Method} &
  \multirow{-2}{*}{Category} &
  \multirow{-2}{*}{\begin{tabular}[c]{@{}c@{}}Crowdedness-aware \\ loss\end{tabular}} &
  \multirow{-2}{*}{\begin{tabular}[c]{@{}c@{}}Confidence and order-aware \\ refinement scheme\end{tabular}} &
  easy &
  medium &
  hard &
  MAE &
  NAE \\ \hline
PSDDN \cite{liu2019point}                  & C-by-D     & - & -          & 60.5      & 60.5      & 39.6      & -        & -                 \\
Shi \textit{et al.} \cite{shi2019counting} & C-by-R     & - & -          & -         & -         & -         & 3.2      & 0.40              \\ \hline
GAK \cite{Zhang2016Single}                 & Size Init. & - & -          & 7.1       & 12.5      & 27.3      & -        & -                 \\
LUDA (ours)                                & Size Init. & - & -          & 7.2       & 12.8      & 29.5      & -        & -                 \\ \hline
                                           &            & \ding{55}  &   \ding{55}         & 18.6      & 23.4      & 30.8      & 3.4      & 0.81              \\
 &
   &
  \cellcolor[HTML]{EFEFEF}\ding{51} &
  \cellcolor[HTML]{EFEFEF}\ding{55} &
  \cellcolor[HTML]{EFEFEF}21.3 &
  \cellcolor[HTML]{EFEFEF}26.8 &
  \cellcolor[HTML]{EFEFEF}35.2 &
  \cellcolor[HTML]{EFEFEF}3.6 &
  \cellcolor[HTML]{EFEFEF}0.97 \\
                                           &            &  \ding{55} & \ding{51} & 55.1      & 55.1      & 52.5      & 2.3      & \textbf{0.27}     \\
\multirow{-4}{*}{Ours} &
  \multirow{-4}{*}{C-by-D} &
  \cellcolor[HTML]{EFEFEF}\ding{51} &
  \cellcolor[HTML]{EFEFEF}\ding{51} &
  \cellcolor[HTML]{EFEFEF}\textbf{75.8} &
  \cellcolor[HTML]{EFEFEF}\textbf{71.0} &
  \cellcolor[HTML]{EFEFEF}\textbf{64.4} &
  \cellcolor[HTML]{EFEFEF}\textbf{2.2} &
  \cellcolor[HTML]{EFEFEF}0.29 \\ \hline
\end{tabular}
\label{tabel2}
\end{table*}

\subsubsection{Evaluation metrics}
For the detection task, we adopted the evaluation protocol in WiderFace \cite{yang2016wider} to calculate average precision (AP). The true positive is defined as the intersection of union (IoU) between ground truth boxes and detected boxes greater than a threshold of \(0.5\). For the counting task, we adopted the commonly-used mean absolute error (MAE) and root mean square error (RMSE) to evaluate the distance between the predicted counts and the ground-truth ones. The MAE indicates the accuracy of methods, while the RMSE reflects their robustness. They are defined as
 
\begin{equation}
    \mathrm{MAE}=\frac{1}{N_{t}}\sum_{N_{t}}^{1}\left | \hat{c}_{i}-c_{i} \right |,  \mathrm{RMSE}=\sqrt{\frac{1}{N_{t}}\sum_{N_{t}}^{1} \left (\hat{c}_{i}-c_{i} \right )^{2}},
\end{equation}  
where \(N_{t}\) is the total number of testing images, and \(\hat{c}_{i}\) and \(c_{i}\) are the estimated count and ground-truth count of the \(i\)-th image, respectively. When counting on WiderFace, we used a normalized MAE (NAE) \cite{shi2019counting}, which normalizes the absolute error by the ground-truth face count.

For the center point localization task, we adopted two evaluation metrics, i.e., AP and mean localization error (MLE) respectively from \cite{liu2019point} and \cite{sam2019locate}. AP defines true positive center point as those whose distance to its ground truth is smaller than a threshold of 20 pixels. MLE calculates the distances in pixels between the predicted center points and the ground truth, and then averages the distances over the testing set. One-to-one matching associates the predictions and the ground truth. The lower the value of MLE is, the better. The AP and MLE are suitable and reasonable for the datasets without bounding-box annotations, e.g., ShanghaiTech. For NWPU-Crowd, we used its evaluation protocol \cite{wang2020nwpu}, i.e., the Precision, Recall, and F1-measure. These metrics evaluate the localization ability of detection-based algorithms in crowded environments. 

\subsection{Ablation Study}
\label{subsec:ExpAbl}
We conducted ablation studies on the WiderFace \cite{yang2016wider} benchmark to evaluate and analyze the improvement of several important modules of our approach, including the locally-uniform distribution assumption (LUDA), the crowdedness-aware loss, and the confidence and order-aware refinement scheme. We evaluated our approach on both the detection and counting tasks. Table \ref{tabel1} shows the AP results of pseudo size generation methods on the training set of WiderFace, while Table \ref{tabel2} shows the AP, MAE, and NAE results on the validation set of WiderFace. The results of state-of-the-art point-supervised detection method (i.e., PSDDN \cite{liu2019point}) and the crowed counting method (i.e., Shi \textit{et al.} \cite{shi2019counting}) are respectively shown in the 1st and 2nd entries of Table \ref{tabel2} for comparison.

\begin{figure} [t]
\centering \includegraphics[width=8.5cm]{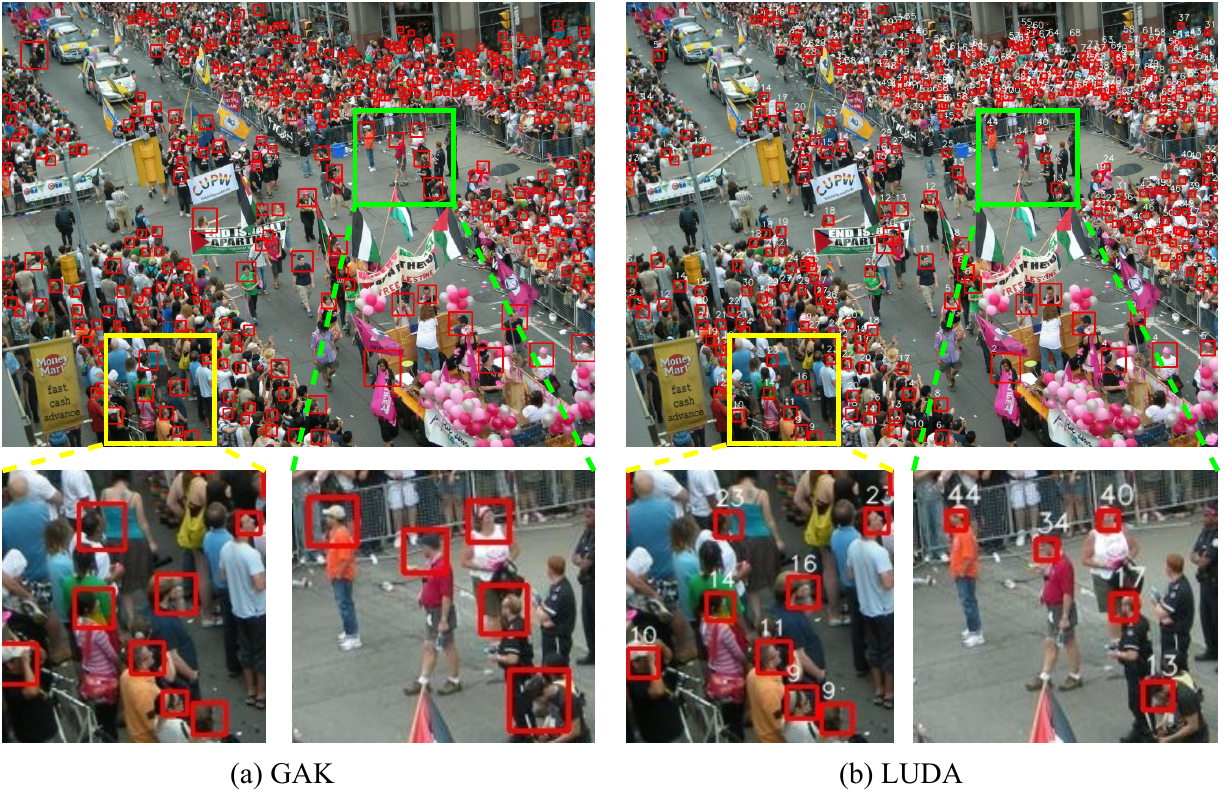}
\caption{Visual comparison of the pseudo object sizes generated by GAK and LUDA in a crowded scene. For LUDA, the number on top of each bounding box refers to the object crowdedness. The high crowdedness value means the pseudo size is accurate.}
\label{fig:size} 
\end{figure}

\subsubsection{Effectiveness of the pseudo object size generation method}
We compared the proposed LUDA-based pseudo size generation method with the geometry-adaptive kernel (GAK) based method in \cite{Zhang2016Single,liu2019point}. The generated boxes were evaluated on the training set of WiderFace. We calculated the AP in three IoU thresholds of \(0.3\), \(0.5\), and \(0.7\), so the true positives become gradually harder to reach. The results are shown in Table \ref{tabel1}, where it can be observed that the proposed LUDA-based method obtains higher AP scores under all thresholds, compared with the GAK-based method, validating the advantage of our LUDA. Such an advantage also means that we can generate more accurate pseudo bounding boxes at the beginning of training. 

Fig. \ref{fig:size} shows a visual comparison of the pseudo object sizes generated by GAK and LUDA in a crowded scene. We can observe that: 1) both methods generate more accurate sizes for crowded objects than sparse objects; and 2) LUDA makes the sparse objects' sizes be accurate when they are surrounded by crowded objects, as illustrated in the green and yellow boxes of Fig. \ref{fig:size}.

We also evaluated the GAK-based and LUDA-based methods by yielding bounding boxes on the validation set of WiderFace. The AP scores are shown in the 3rd and 4th entries of Table \ref{tabel2}, where it can be seen that the accuracy of the generated bounding boxes is extremely low, and especially the AP score is only 7.2\% on the easy subset. This observation shows that only applying pseudo size generation methods cannot obtain accurate object sizes even with the ground-truth center points.

\subsubsection{Effectiveness of the crowdedness-aware loss and the confidence and order-aware refinement scheme}
The crowdedness-aware loss and the confidence and order-aware refinement scheme are the critical components to improve the AP for point-supervised detection. Based on the pseudo bounding boxes, we trained the detector on the training set of WiderFace with four sittings 1) using only the crowdedness-aware loss, 2) using only the confidence and order-aware refinement scheme, 3) using neither of the two modules, and 4) using both of the two modules. The AP, MAE, and NAE results are listed in the 5th to 8th entries of Table \ref{tabel2}. 

\textbf{Pseudo sizes only.} The detector was trained with the pseudo object sizes only, as listed in the 5th entry of Table \ref{tabel2}. The AP scores are improved by about 11\% on the easy and medium subsets in comparison with the LUDA-based size generation method. However, the values are still low, e.g., 18.6\%, 23.4\%, and 30.8\% AP on the easy, medium, and hard subsets, respectively. This observation demonstrates that it is hard to attain an acceptable detector when trained only with pseudo bounding boxes. 

\textbf{Either the crowdedness-aware loss or the confidence and order-aware refinement scheme.} As shown in the 6th and 7th entries of Table \ref{tabel2}, with the use of the crowdedness-aware loss (resp. the confidence and order-aware refinement scheme), the AP scores increase to 21.3\% (easy), 26.8\% (medium), and 30.8\% (hard) (resp. 55.1\% (easy), 55.1\% (medium), and 52.5\% (hard)), which validate the effectiveness of these two modules. Moreover, it can be known that the refinement module is more effective than the loss. Note that only with the refinement scheme, our approach outperforms PSDDN \cite{liu2019point} on the hard subset in the detection task (52.5\% AP vs. 39.4\% AP) and Shi \textit{et al.} \cite{shi2019counting} in the counting task (2.3 MAE vs. 3.2 MAE). 

\textbf{Both the crowdedness-aware loss and the confidence and order-aware refinement scheme.} When both modules are activated, our approach achieves the best results in both detection and counting tasks. There is a big jump of AP from using a single module to both modules. We think that the crowdedness-aware loss emphasizes the accurate pseudo sizes, but neglects to update the noisy sizes during training, while the confidence and order-aware refinement scheme updates the pseudo sizes but it overlooks the importance of the accurate pseudo sizes in the network's weights updating. The combination of such two modules could well compensate to each other. Numerically, the AP scores of our approach with both two modules  increase to 75.8\% (easy), 71.0\% (medium), and 64.4\% (hard). Compared with PSDDN \cite{liu2019point}, our method improves the AP by more than 10\%. The MAE and NAE of our approach respectively decrease approximately 31.2\% and 27.5\% against \cite{shi2019counting}. The results demonstrate the significant superiority of our method in both detection and counting tasks.

\begin{figure} [t]
\centering \includegraphics[width=8.5cm]{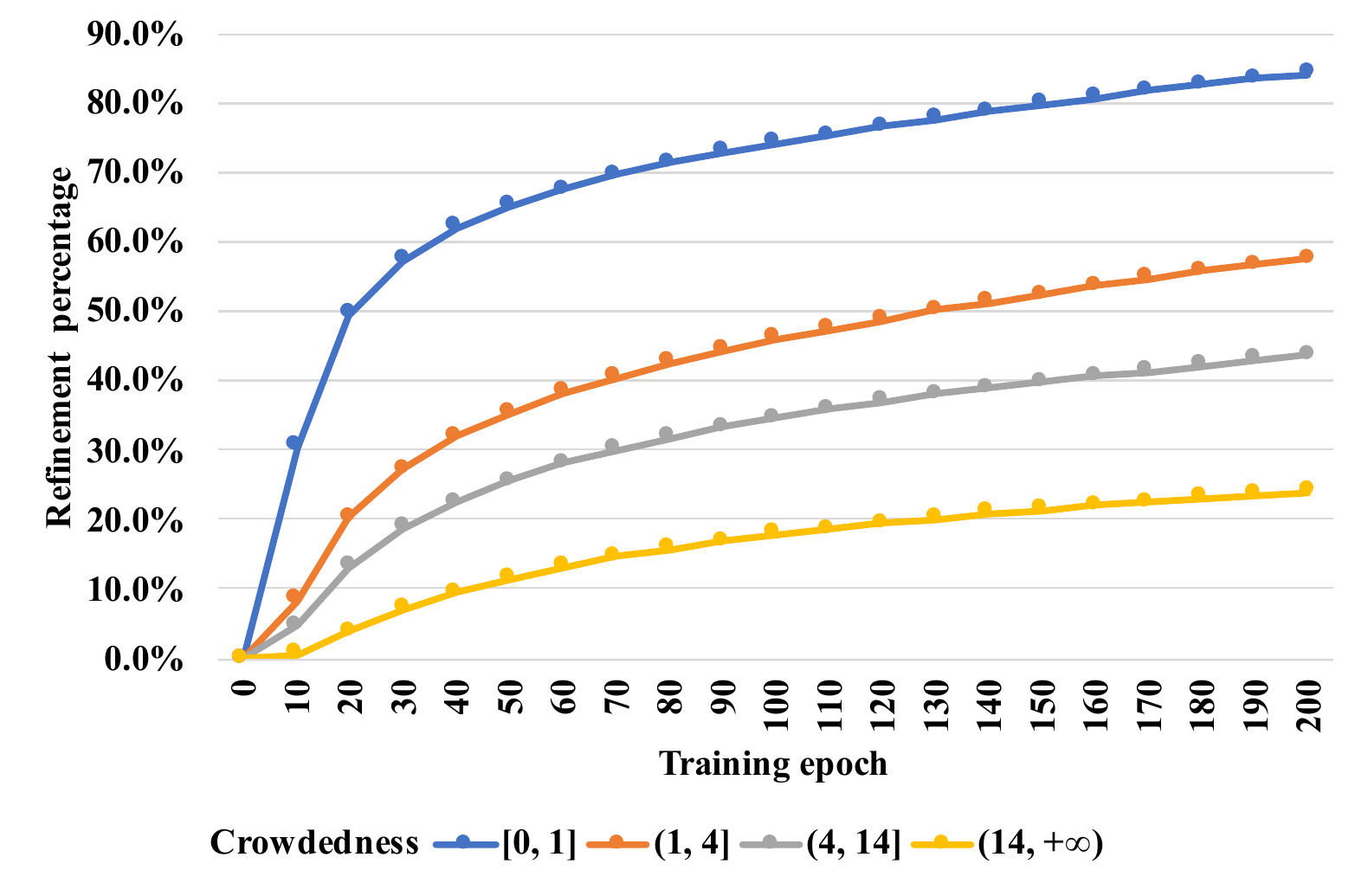}
\caption{The percentage of refined training examples with respect to the training epoch.}
\label{fig:example} 
\end{figure}

\begin{figure*} [t]
\centering \includegraphics[width=\textwidth]{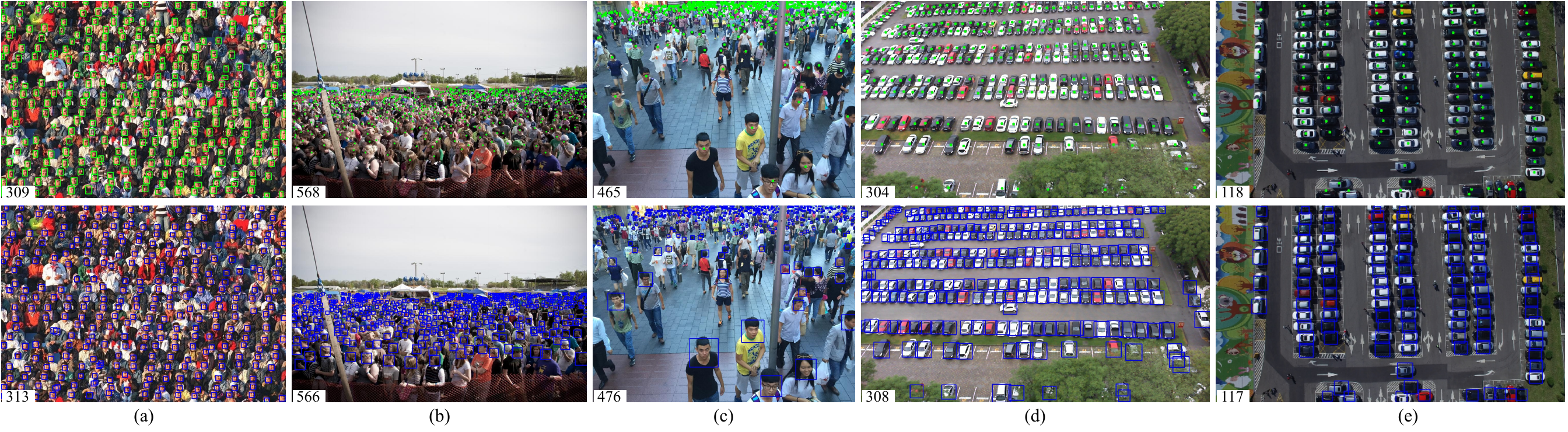}
\caption{Qualitative results on (a) WiderFace \cite{yang2016wider}, (b) SHA \cite{Zhang2016Single}, (c) SHB \cite{Zhang2016Single}, (d) PUCPR+ \cite{hsieh2017drone}, and (e) CARPK \cite{hsieh2017drone}. The top row shows the ground-truth boxes or points, and counts. The bottom row shows the bounding boxes and counts predicted by our approach. \textit{Zoom in the figure for better viewing}.}
\label{fig3} 
\end{figure*}

Besides, the last row of Table \ref{tabel1} shows that the refined pseudo sizes improve the AP (0.3) and AP (0.5) by nearly 20\%. The gradually enhanced quality of training examples helps the detector become stronger. To demonstrate this improvement, we plot some training examples before and after training on the WiderFace dataset in Fig. \ref{fig1}. Especially for large and sparse objects, the bounding boxes are refined to encompass the face regions.

\subsubsection{Behavior of the confidence and order-aware refinement scheme}
In our confidence and order-aware refinement scheme, the pseudo object sizes are updated in a robust and orderly manner, i.e., the easy examples (sparse and large objects) with inaccurate sizes are first refined, followed by the hard examples (crowded and small objects). The object sizes are refined when their posterior probabilities are larger than the threshold of 0.6. To illustrate the behavior of such a refinement scheme, we counted the percentage of examples whose probabilities exceed 0.6 during training. Specifically, we split examples into four categories corresponding to the crowdedness intervals $[0, 1]$, $(1, 4]$, $(4, 14]$, and $[14,+\infty)$ with 64k, 30k, 31k, and 30k training examples, respectively. Fig. \ref{fig:example} shows the percentage of examples with their posterior probabilities larger than 0.6 under each category. We can observe that the percentage of the revised low-crowdedness examples increases more rapidly than that of the high-crowdedness examples in the first 20 epochs. Throughout the entire training epochs, the percentage corresponding to lower-crowdedness reaches the highest, demonstrating the effectiveness of our refinement scheme.

\begin{table}[t]
\centering \caption{Comparisons with state-of-the-art point-supervised detection methods on the validation set of WiderFace \cite{yang2016wider}. ``*": the AP results are provided by \cite{liu2019point}. The larger the value of AP is, the better. The initialized pseudo boxes were used when training CSP with point supervision.}
\begin{tabular}{l|c|ccc}
\hline
\hline
Method       & Supervision & easy          & medium        & hard          \\ \hline
LSC-CNN \cite{sam2019locate}    & Box       & 57.3      & 70.1      & 68.9    \\ 
Faceness-WIDER \cite{yang2015facial} & Box         & 71.3          & 63.4          & 34.5          \\ 
Faster R-CNN* \cite{ren2015faster} & Box         & 84.0          & 72.4          & 34.7          \\
CSP (anchor-free) \cite{Liu_2019_CVPR}   & Box  &  90.7  &  \textbf{95.2}  &  \textbf{96.1}  \\
HR-TinyFace \cite{hu2017finding}  & Box         & \textbf{92.5}          & 91.0          & 80.6          \\ 
\hline \hline
CSP (anchor-free) \cite{Liu_2019_CVPR}   & Point  &  18.6  &  23.4  &  30.8 \\
CSR-A-thr \cite{sam2019locate}  & Point       & 30.2          & 41.9          & 33.5 \\
PSDNN \cite{liu2019point} & Point       & 60.5          & 60.5          & 39.6          \\
LSC-CNN \cite{sam2019locate}    & Point       & 40.5      & 62.1      & 46.2    \\
Ours         & Point       & \textbf{75.8} & \textbf{71.0} & \textbf{64.4} \\ \hline
\end{tabular}
\label{tabel3}
\end{table}

\subsection{Comparison with State-of-the-Art Methods}
\label{subsec:Expcomp}
\subsubsection{Point-supervised face detection}
We compared our approach with state-of-the-art point-supervised detection methods, including PSDDN \cite{liu2019point} and LSC-CNN \cite{sam2019locate}, and the counting-by-regression method, i.e., CSR-A-thr \cite{sam2019locate}. The CSR-A-thr is the detection version of CSRNet \cite{li2018csrnet}. Table \ref{tabel3} shows the AP scores of different methods on the validation set of WiderFace \cite{yang2016wider}. From Table \ref{tabel3}, we can see that our method outperforms the other point-supervised methods to a significant margin (i.e., more than 10\% AP improvement). Especially on the hard subset, our method improves the AP score of the second best LSC-CNN method by 18.2\%. 

\begin{figure*} [t]
\centering \includegraphics[width=\textwidth]{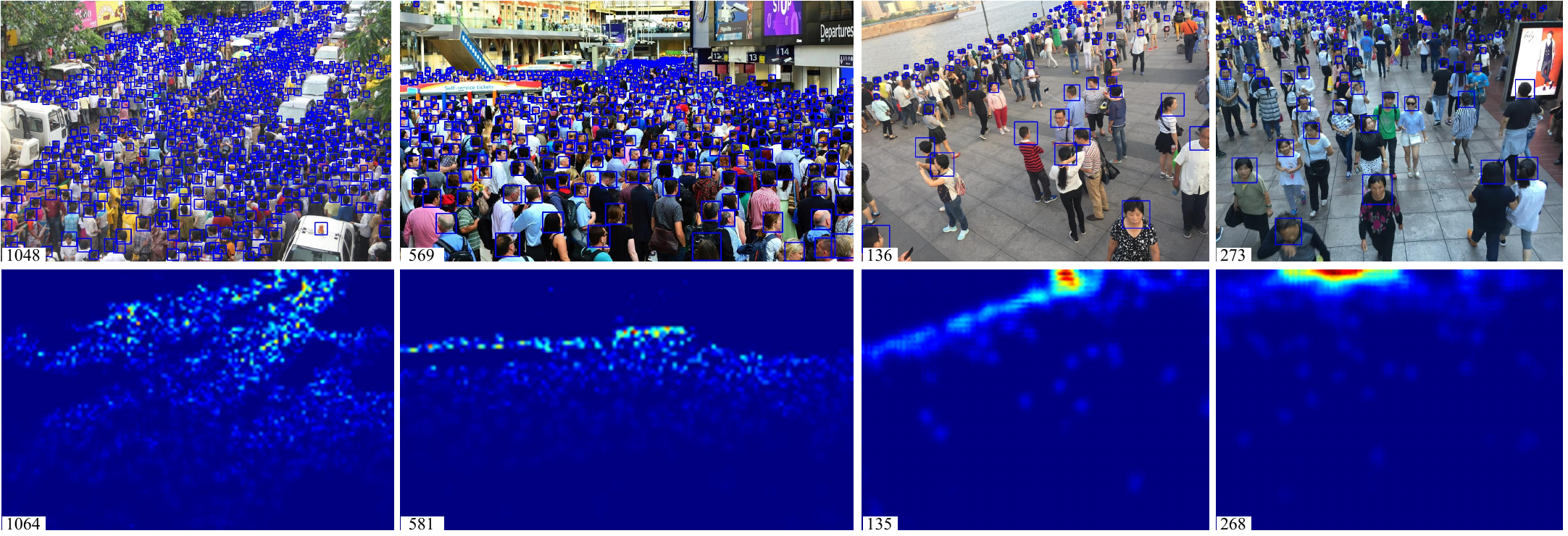}
\caption{Comparisons between the bounding boxes (at top row) produced by our approach and the density maps (at bottom row) produced by CSRNet \cite{li2018csrnet} on the SHA and SHB \cite{Zhang2016Single} datasets. The predicted count is shown at the bottom-left corner of an image. The ground-truth counts from left to right images are 1068, 584, 139, and 274. \textit{Zoom in the figure for better viewing}.}
\label{fig4} 
\end{figure*}

Besides, Table \ref{tabel3} provides the results of several box-supervised methods for comparison, where it can be seen that our method even outperforms box-supervised LSC-CNN and Faceness-WIDER on the easy subset and Faceness-WIDER and Faster-RCNN on the hard subset. Although the AP score of the box-supervised CSP achieves above 90\%, it significantly drops to 30.8\% when trained with point-initialized pseudo boxes, which demonstrates the effectiveness of our point-supervised training scheme. Fig. \ref{fig3} shows some visual results of bounding box predicted by our method.

\begin{table}[t]
\centering \caption{Comparisons with crowd localization methods on SHA and SHB \cite{Zhang2016Single} datasets. The larger the value of AP is and the lower the value of MLE is, the better.}
\begin{tabular}{l|cc|cc}
\hline \hline
\multirow{2}{*}{Method}       & \multicolumn{2}{c|}{SHA} & \multicolumn{2}{c}{SHB} \\ \cline{2-5} 
                              & AP (\%)      & MLE        & AP (\%)      & MLE        \\ \hline
PSDDN \cite{liu2019point}     & 73.7        & -          & 75.9        & -          \\
CSR-A-thr \cite{sam2019locate} & -           & 16.8       & -           & 12.3       \\
LSC-CNN \cite{sam2019locate}  &   67.6          & 9.6        &     76.0        & 9.0        \\
Ours                          & \textbf{85.3}  & \textbf{8.0} & \textbf{91.6}  & \textbf{6.0} \\ \hline
\end{tabular}
\label{tabel5}
\end{table}

\subsubsection{Center point localization of crowds} 
\textbf{SHA and SHB datasets.} To evaluate the localization ability for the datasets with only point-level annotations, we compared our method with PSDNN \cite{liu2019point}, CSR-A-thr \cite{sam2019locate}, and LSC-CNN \cite{sam2019locate} on the SHA and SHB \cite{Zhang2016Single} datasets. Table \ref{tabel5} presents the results of the AP and MLE metrics. Our approach obtains the best results, showing 85.3\% AP and 8.0 MLE for SHA, and 91.6\% AP and 6.0 MLE for SHB. 

\begin{table}[t]
\caption{Comparison with state-of-the-art crowd localization methods on the NWPU-Crowd test set. The values of F1-measure, Precision, and Recall were calculated under the threshold $\sigma_l$ \cite{wang2020nwpu}.}
\label{tab:localization}
\resizebox{\columnwidth}{!}{%
\begin{tabular}{l|c|c|c|c|c}
\hline
Method      & Supervision & Output & F1  & Precision  & Recall  \\
&  & &(\%) & (\%)& (\%)\\\hline
Faster R-CNN \cite{ren2015faster} & Box   & Box          & 6.7       & \textbf{95.8}     & 3.5      \\ \hline
TinyFaces \cite{hu2017finding}   & Box  & Box            & 56.7      & 52.9     & 61.1     \\ \hline\hline
VGG+GPR \cite{domainadaptive}     & Point  & Point           & 52.5      & 55.8     & 49.6     \\ \hline
RAZ\_Loc \cite{RAZ_Loc}     & Point  & Point         & 59.8      & 66.6     & 54.3     \\ \hline
Ours        & Point  & Box &  \textbf{63.7}       &  65.1      &  \textbf{62.4}      \\ \hline
\end{tabular}
}
\end{table}

\textbf{NWPU-Crowd dataset.} We also compared our approach with state-of-the-art crowd localization methods on the large-scale NWPU-Crowd \cite{wang2020nwpu} dataset. The results are shown in Table \ref{tab:localization}, where it can be observed that our method obtains the best F1-measure (i.e., 63.7\%) and Recall (i.e., 62.4\%). Fast R-CNN achieves 95.8\% Precision but sacrifices the Recall to 3.5\%, indicating that Fast R-CNN fails to detect crowded people. Note that the proposed approach is the only one that can output bounding boxes when trained with point supervision.

\begin{table}[t]
\centering \caption{Comparisons with state-of-the-art counting-by-regression methods (in the top part) and counting-by-detection methods (in the bottom part) on SHA and SHB \cite{Zhang2016Single}, and WiderFace \cite{yang2016wider} (WF) datasets. ``-": the author does not provide the result. ``*": the results are provided by \cite{sam2019locate}. The lower the values of MAE and RMSE are, the better.}
\resizebox{\columnwidth}{!}{%
\begin{tabular}{l|cc|cc|c}
\hline \hline
\multirow{2}{*}{Method} & \multicolumn{2}{c|}{SHA}       & \multicolumn{2}{c|}{SHB}     & WF           \\ \cline{2-6} 
                        & MAE           & RMSE            & MAE          & RMSE           & MAE          \\ \hline
Zhang \textit{et al.} \cite{Zhang2016Single}            & 110.2         & 173.2          & 26.4         & 41.3          & 7.1          \\
CSRNet \cite{li2018csrnet}              & 68.2          & 115.0          & 10.6         & 16.0          & 4.3          \\
Cao et al. \cite{cao2018scale}             & 67.0          & 104.5          & 8.4          & 13.6          & 8.5          \\
PSDNN+ \cite{liu2019point}                 & 65.9          & 112.3          & 9.1          & 14.2          & -            \\
Shi \textit{et al.} \cite{shi2019counting}              & 65.2 & 109.4 & \textbf{7.2} & \textbf{12.2} & \textbf{3.2} \\
PSD+DCL \cite{DensityAware}  & 65.0 & 108.0 & 8.1 & 13.3 & - \\
HA-CCN \cite{sindagi2019ha}   & \textbf{62.9} & \textbf{94.9} & 8.1 & 13.4 & \textbf{-} \\ 
\hline \hline
TinyFace* \cite{hu2017finding}                & 237.8         & 422.8          & -            & -             & -          \\
LC-FCN8 \cite{laradji2018blobs}                 & -             & -              & 13.1         & -             & -            \\
PSDNN \cite{liu2019point}                   & 85.4          & 159.2          & 16.1         & 27.9          & -            \\
LSC-CNN \cite{sam2019locate}                 & 66.4          & 117.0          & 8.1         & 15.7          & -            \\
Ours                    & \textbf{65.1}           & \textbf{104.4}            & \textbf{7.8}          & \textbf{12.6}           & \textbf{2.2} \\ \hline
\end{tabular}
}
\label{tabel4}
\end{table}

\subsubsection{Crowd counting} 
In addition to detection, we evaluated our method in the crowd counting task. Table \ref{tabel4} shows the results on SHA and SHB \cite{Zhang2016Single}, and WiderFace \cite{yang2016wider} datasets. For a fair comparison, we split the counting methods into two categories: counting-by-regression methods and counting-by-detection methods. Our method achieves the best performance when compared with state-of-the-art counting-by-detection methods. For SHA and SHB, the proposed method is comparable to state-of-the-art counting-by-regression methods, such as Shi \textit{et al.} \cite{shi2019counting} and HA-CCN \cite{sindagi2019ha}. 

Note that our method not only provides the count but also estimates the bounding boxes for object instances. We qualitatively evaluated the proposed method by visualizing the predicted bounding boxes on the SHA, and SHB datasets in Fig. \ref{fig3}. Besides, Fig. \ref{fig4} shows the predicted boxes by our method and the estimated density maps by CSRNet \cite{li2018csrnet} on the SHA and SHB datasets. We argue that the box outputs of our method are more informative than the density maps of counting-by-regression methods because the boxes provide high-level understanding of crowds.

\begin{table}[t]
\centering \caption{Comparisons with state-of-the-art vehicle counting methods on the CARPK and PUCPR+ benchmark \cite{hsieh2017drone}. The lower the values of MAE and RMSE are, the better.}
\begin{tabular}{l|cc|cc}
\hline
\hline
\multirow{2}{*}{Method} & \multicolumn{2}{c|}{CARPK} & \multicolumn{2}{c}{PUCPR+} \\ \cline{2-5} 
                        & MAE          & RMSE         & MAE          & RMSE         \\ \hline
LPN Counting \cite{hsieh2017drone}            & 23.80        & 36.79       & 22.76        & 34.46       \\
RetinaNet \cite{Lin_2017_ICCV}               & 16.62        & 22.30       & 24.58        & 33.12       \\
IEP Counting \cite{stahl2018divide}            & 51.83        & -           & 15.17        & -           \\
Goldman \textit{et al.} \cite{goldman2019precise}          & 6.77         & 8.52        & 7.16         & 12.00       \\
Li \textit{et al.} \cite{li2019simultaneously}          & 5.24        & 7.38        & 3.92         & 5.06       \\
Ours                    & \textbf{4.95}        & \textbf{7.09}         & \textbf{3.20}          & \textbf{4.83}         \\ \hline
\end{tabular}
\label{tabel7}
\end{table}

\subsubsection{Vehicle counting from drone view}
To evaluate the generalization ability in other domains, we trained our model on the CARPK and PUCPR+ datasets \cite{hsieh2017drone}. The datasets provide densely-packed car counting images from drone view. Table \ref{tabel7} reports the MAE and RMSE values of our method and state-of-the-art vehicle counting methods \cite{hsieh2017drone,Lin_2017_ICCV,stahl2018divide,goldman2019precise,li2019simultaneously}. It can be seen that our method consistently performs better than the other methods, demonstrating that our model is flexible for various detection and counting tasks. Some visual results on CARPK and PUCPR+ are shown in Fig. \ref{fig3}.

\section{Conclusion}
\label{Sec:Con} 
In this paper, we have presented a self-training approach to train a typical detector with only point-level annotations such that the detector can accurately detect and count objects in crowd scenes simultaneously. This is achieved by the locally-uniform distribution assumption, the crowdedness-aware loss, the confidence and order-aware refinement scheme, and the decoding method to promote the detector to generate accurate bounding boxes in a coarse-to-fine and end-to-end manner. Extensive experimental results over multiple commonly used benchmark datasets have demonstrated that the proposed approach achieves the best performance in point-supervised detection and counting tasks among detection-based methods, and our method even produces comparable performance to state-of-the-art counting-by-regression methods. We believe that DNN-based object detection in crowds with only point supervision is a potential and promising research area.


%

\ifCLASSOPTIONcaptionsoff
  \newpage
\fi



%

%
\bibliographystyle{IEEEtran}
\bibliography{myref}

\begin{IEEEbiography}[{\includegraphics[width=1in,height=1.25in,clip,keepaspectratio]{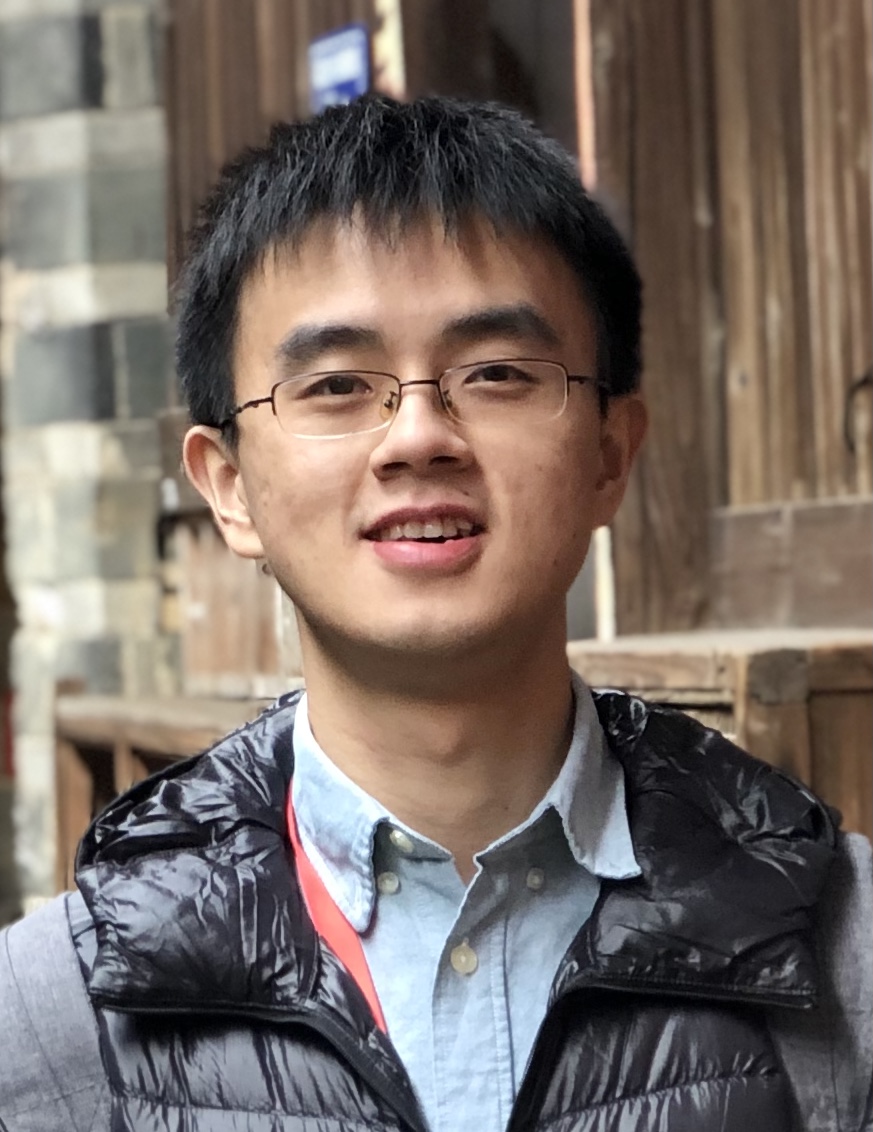}}]{Yi Wang}
received the B.Eng. degree in electronic information engineering and M.Eng. degree in information and signal processing from the School of Electronics and Information, Northwestern Polytechnical University, Xi'an, China, in 2013 and 2016, respectively. He is currently a research associate with the School of Electrical and Electronic Engineering, Nanyang Technological University, Singapore, and he is also working toward the Ph.D. degree of Nanyang Technological University. His research interests include image restoration, image recognition, object detection, and crowd analysis.
\end{IEEEbiography}

\begin{IEEEbiography}[{\includegraphics[width=1in,height=1.25in,clip,keepaspectratio]{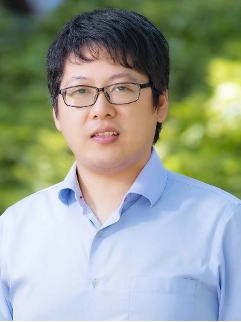}}]{Junhui Hou}
received the B.Eng. degree in information engineering (Talented Students Program) from the South China University of Technology, Guangzhou, China, in 2009, the M.Eng. degree in signal and information processing from Northwestern Polytechnical University, Xi'an, China, in 2012, and the Ph.D. degree in electrical and electronic engineering from the School of Electrical and Electronic Engineering, Nanyang Technological University, Singapore, in 2016. He has been an Assistant Professor with the Department of Computer Science, City University of Hong Kong, since 2017. His research interests fall into the general areas of visual computing, such as image/video/3D geometry data representation, processing and analysis, semi/un-supervised data modeling, and data compression and adaptive transmission

Dr. Hou was the recipient of several prestigious awards, including the Chinese Government Award for Outstanding Students Study Abroad from China Scholarship Council in 2015, and the Early Career Award (3/381) from the Hong Kong Research Grants Council in 2018. He is a member of MSA-TC and VSPC-TC, IEEE CAS. He is currently serving as an Associate Editor for IEEE Transactions on Circuits and Systems for Video Technology, The Visual Computer, and Signal Processing: Image Communication, and the Guest Editor for the IEEE Journal of Selected Topics in Applied Earth Observations and Remote Sensing. He also served as an Area Chair of ACM MM 2019 and 2020, IEEE ICME 2020, and WACV 2021. He is a senior member of IEEE.
\end{IEEEbiography}

\begin{IEEEbiography}[{\includegraphics[width=1in,height=1.25in,clip,keepaspectratio]{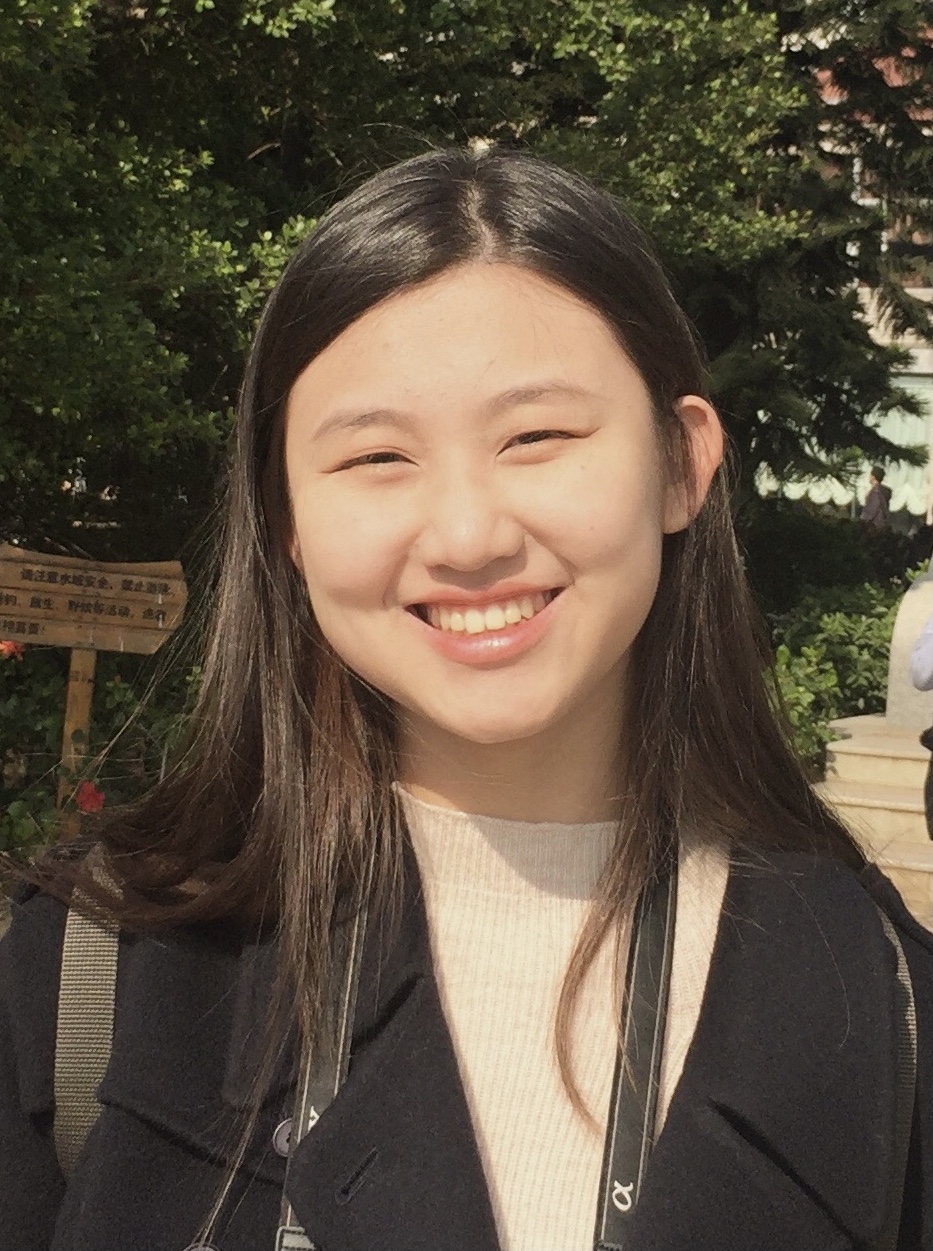}}]{Xinyu Hou}
received the B.Eng. degree from the School of Electrical and Electronic Engineering, Nanyang Technological University, Singapore, in 2020. She is currently a project officer at the School of Computer Science and Engineering, Nanyang Technological University. Her research interests include image/video generation, computer vision, and machine learning.
\end{IEEEbiography}

\begin{IEEEbiography}[{\includegraphics[width=1in,height=1.25in,clip,keepaspectratio]{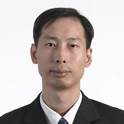}}]{Lap-Pui Chau}
received the Bachelor degree from Oxford Brookes University,  and the Ph.D. degree from The Hong Kong Polytechnic University, in 1992 and 1997, respectively.  He is Assistant Chair (Academic) of School of Electrical and Electronic Engineering, Nanyang Technological University. His research interests include visual signal processing algorithms, light-field imaging, video analytics for intelligent transportation system, and human motion analysis.
 
He was a General Chairs for IEEE International Conference on Digital Signal Processing (DSP 2015) and International Conference on Information, Communications and Signal Processing (ICICS 2015). He was a Program Chairs for Visual Communications and Image Processing (VCIP 2020 and VCIP 2013), International Conference on Digital Signal Processing (DSP 2018), International Conference on Multimedia and Expo (ICME 2016) and International Symposium on Intelligent Signal Processing and Communications Systems (ISPACS 2010).
 
He was the chair of Technical Committee on Circuits \& Systems for Communications (TC-CASC) of IEEE Circuits and Systems Society from 2010 to 2012. He served as an associate editor for IEEE Transactions on Multimedia, IEEE Signal Processing Letters, IEEE Transactions on Circuits and Systems for Video Technology, IEEE Transactions on Circuits and Systems II, and is currently serving as an associate editor for IEEE Transactions on Broadcasting, and The Visual Computer (Springer Journal). Besides, he was an IEEE Distinguished Lecturer for 2009-2019, and a steering committee member of IEEE Transactions for Mobile Computing from 2011-2013. He is an IEEE Fellow.
\end{IEEEbiography}




\end{document}